\documentclass[10pt,twocolumn,letterpaper]{article}

\usepackage{cvpr}              

\usepackage[accsupp]{axessibility}
\usepackage{graphicx}
\usepackage{amsmath}
\usepackage{amssymb}
\usepackage{booktabs}

\usepackage{wrapfig}
\usepackage{enumerate}
\usepackage{enumitem}
\usepackage{tabularx}
\usepackage{longtable}
\usepackage{ltablex}
\usepackage{makecell}

\usepackage[pagebackref,breaklinks,colorlinks]{hyperref}

\usepackage{xcolor}

\usepackage[capitalize]{cleveref}
\crefname{section}{Sec.}{Secs.}
\Crefname{section}{Section}{Sections}
\Crefname{table}{Table}{Tables}
\crefname{table}{Tab.}{Tabs.}
\Crefname{figure}{Figure}{Figures}
\crefname{figure}{Fig.}{Figs.}

\usepackage{xspace}
\makeatletter
\DeclareRobustCommand\onedot{\futurelet\@let@token\@onedot}
\def\@onedot{\ifx\@let@token.\else.\null\fi\xspace}

\def\eg{\emph{e.g}\onedot} 
\def\ie{\emph{i.e}\onedot}

\makeatother

\def\cvprPaperID{2115} 
\def\confName{CVPR}
\def\confYear{2023}

\begin{document}

\title{Super-CLEVR: A Virtual Benchmark to \\Diagnose Domain Robustness in Visual Reasoning}


\author{%
Zhuowan Li\textsuperscript{$1$} \qquad Xingrui Wang\textsuperscript{$2$}  \qquad Elias Stengel-Eskin \textsuperscript{$1$} \\ 
\enspace Adam Kortylewski\textsuperscript{$3,4$} \qquad Wufei Ma\textsuperscript{$1$} \qquad Benjamin Van Durme\textsuperscript{$1$} \qquad Alan Yuille\textsuperscript{$1$} \\ 
{\textsuperscript{$1$} Johns Hopkins University    } \qquad
{\textsuperscript{$2$} University of Southern California} \qquad \\
{\textsuperscript{$3$} Max Planck Institute for Informatics} \qquad {\textsuperscript{$4$} University of Freiburg} \quad \\
}

\maketitle


\begin{abstract}
  \textit{Visual Question Answering (VQA)} models often perform poorly on out-of-distribution data and struggle on domain generalization. Due to the multi-modal nature of this task, multiple factors of variation are intertwined, making generalization difficult to analyze.
  This motivates us to introduce a virtual benchmark, Super-CLEVR, where different factors in VQA domain shifts can be isolated in order that their effects can be studied independently. Four factors are considered: visual complexity, question redundancy, concept distribution and concept compositionality. With controllably generated data, Super-CLEVR enables us to test VQA methods in situations where the test data differs from the training data along each of these axes. We study four existing methods, including two neural symbolic methods NSCL\cite{Mao2019NeuroSymbolic} and NSVQA\cite{nsvqa}, and two non-symbolic methods FiLM \cite{perez2018film} and mDETR\cite{kamath2021mdetr}; and our proposed method, probabilistic NSVQA (P-NSVQA), which extends NSVQA with uncertainty reasoning. P-NSVQA outperforms other methods on three of the four domain shift factors. Our results suggest that disentangling reasoning and perception, combined with probabilistic uncertainty, form a strong VQA model that is more robust to domain shifts. 
  The dataset and code are released at \url{https://github.com/Lizw14/Super-CLEVR}.

\end{abstract}

\section{Introduction}


Visual question answering (VQA) is a challenging task that assesses the reasoning ability of models to answer questions based on both visual and linguistic inputs. 
Current VQA methods are typically developed on standard benchmarks like VQAv2 \cite{balanced_vqa_v2} or GQA \cite{hudson2019gqa}, with the implicit assumption that testing data comes from the same underlying distribution as training data. However, as has been widely studied in computer vision \cite{ganin2016domain, qiao2020learning, li2019episodic}, algorithms trained on one domain often fail to generalize to other domains. 
Moreover, having learned the distributional prior of training data,  models often struggle on out-of-distribution tests. This has been  studied in VQA from the perspective of domain transfer \cite{chao2018cross, zhang2021domain, xu2019open}, dataset bias \cite{vqa-cp, Niu_2021_CVPR, Dancette2021BeyondQB}, counter-factual diagnosis \cite{Niu2021CounterfactualVA, CSS}, and out-of-distribution benchmarking \cite{GQA-OOD}.

The multi-modal nature of VQA gives rise to multiple intertwined factors of variation, making domain shift an especially difficult problem to study.
For example, \cite{chao2018cross} suggests that VQA domain shifts are a combination of differences in images, questions or answers; and \cite{Li_2021_ICCV} reveals a gap between synthetic and real VQA datasets by differences in the over-specification of questions and the underlying distribution of concepts. However, despite a wealth of research on domain generalization in VQA \cite{akula2021crossvqa, jiang2021x, xu2019open, zhang2021domain}, there is no systematic analysis of the contributing factors in domain shifts. 

\begin{figure*}[t!]
\begin{center}
    \vspace{-1.9em}
    \includegraphics[width=0.98 \linewidth]{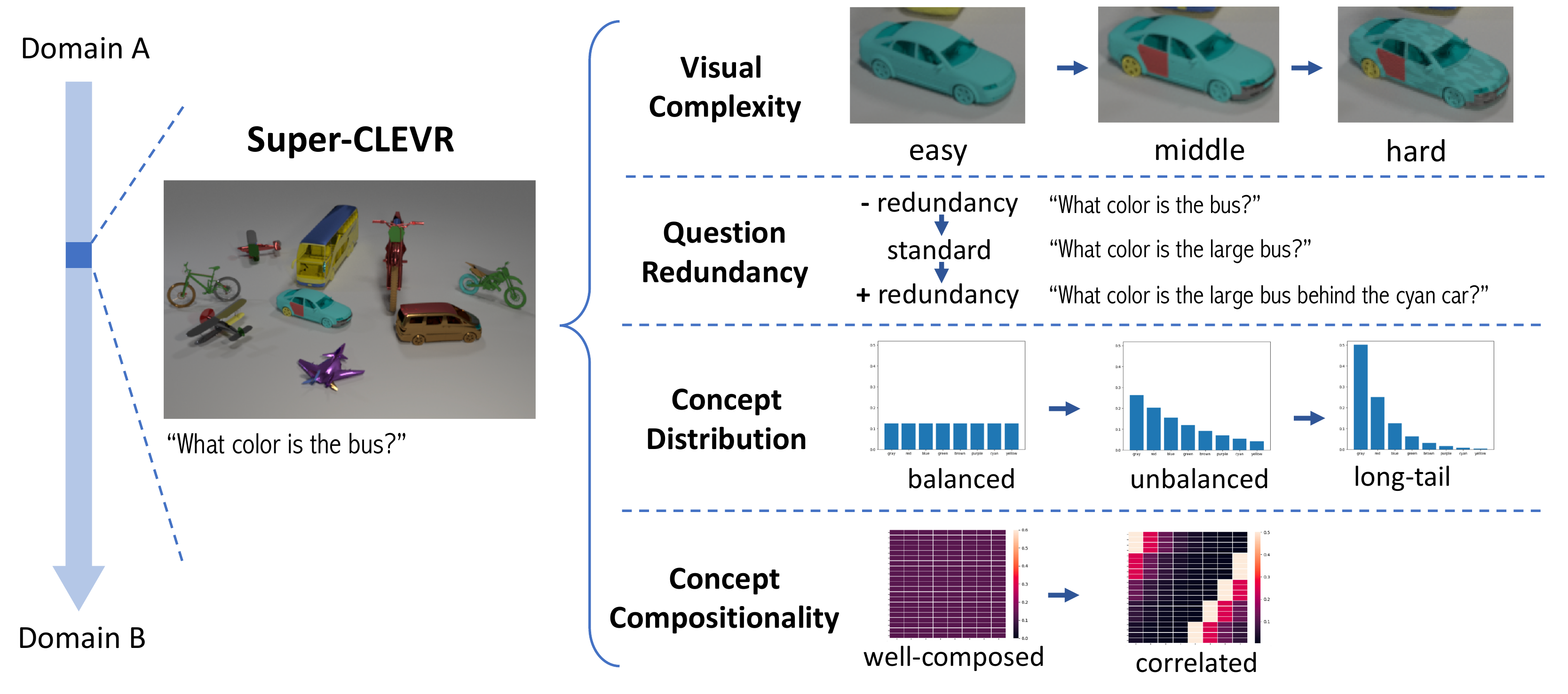}
    \vspace{-0.9em}
    \caption{We decompose VQA domain shifts into four contributing factors: visual complexity, question redundancy, concept distribution and concept compositionality. The domain shifts along each factor can be independently studied with the proposed Super-CLEVR dataset. }
    \vspace{-2.0em}
    \label{fig:introduction}
\end{center}
\end{figure*}

To this end, we introduce a virtual benchmark, Super-CLEVR, which enables us to test VQA algorithms in situations where the test data differs from the training data. We decompose the domain shift into a set of isolated contributing factors, so that their effects can be diagnosed independently. We study four factors: visual complexity, question redundancy, concept distribution, and concept compositionality. These are illustrated in \cref{fig:introduction} and described in \cref{sec: motivation}. With controllable data generation using our SuperCLEVR virtual benchmark, we are able to isolate the different factors in VQA domain shifts so that their effects can be studied independently. Compared with the original CLEVR dataset \cite{johnson2017clevr}, Super-CLEVR contains more complicated visual components and has better controllability over the domain shift factors. As shown in \cref{fig:introduction}, the Super-CLEVR dataset contains images rendered from 3D graphical vehicle models in the UDA-Part dataset \cite{liu2022learning}, paired with questions and answers automatically generated from templates. The objects and questions are sampled based on the specified underlying probability distribution, which can be controlled to produce distribution shifts in different factors.


With Super-CLEVR, we diagnose the domain robustness of current VQA models. Four representative models are studied: for the classic two-stream feature fusing architecture, we choose FiLM \cite{perez2018film}; for a large-scale pretrained model we take mDETR \cite{kamath2021mdetr}; we use NSCL \cite{Mao2019NeuroSymbolic} and NSVQA \cite{nsvqa} as representative neuro-symbolic methods. 
We observe that all these models suffer from domain shifts to varying degrees of sensitivity. We analyze each factor separately to examine the influence of different model designs. 
Specifically, we find that the step-by-step design of neural modular methods enhances their robustness to changes in question redundancy compared with non-modular ones; however, the non-modular models are more robust to visual complexity. Furthermore, thanks to its decomposed reasoning and perception, NSVQA is more robust to concept distribution shifts.

While existing models suffer from domain shifts with different characteristics, we make a technical improvement over NSVQA which enables it to significantly outperform existing models on three of the four factors. In particular, we inject probabilities into the deterministic symbolic executor of NSVQA, empowering it to take into account the uncertainty of scene understanding. We name our model \emph{probabilistic NSVQA} (P-NSVQA), and show that its performance improvement in both the in-domain and out-of-domain settings. With superior results of P-NSVQA, we suggest that disentangling reasoning from vision and language understanding, together with probabilistic uncertainty, gives a strong model that is robust to domain shifts.


Our contributions are as follows. (1) We introduce the Super-CLEVR benchmark to diagnose VQA robustness along four different factors independently. This benchmark can also be used for part-based reasoning. (2) We enhance a neural-symbolic method by taking the uncertainty of visual understanding into account in reasoning. (3) We conduct detailed analysis of four existing methods, as well as our novel approach to study the influence of model designs on distinct robustness factors. We conclude that  disentangled reasoning and perception plus explicit modeling of uncertainty leads to a more robust VQA model.

\section{Related work}
\textbf{Visual question answering (VQA).}
Popular VQA methods fall into three categories. Two-stream methods extract features for image and questions using CNN and LSTM respectively, then enable interaction between the two modalities with different feature fusing methods \cite{anderson2018bottom, fukui2016multimodal, kim2018bilinear, yu2019mcan, hudson2018compositional, perez2018film, li2019relation}.
Neural symbolic methods, on the other hand, use a parse-then-execute pipeline where the question is parsed into a functional program, which is then executed on the image using neural modules \cite{nsvqa, Mao2019NeuroSymbolic}. 
Recently, transformers-based models have achieved impressive performance on various vision-and-language tasks by pretraining on large scale dataset then finetuning for downstream tasks \cite{tan2019lxmert, lu2019vilbert, li2020oscar, zhang2021vinvl, kamath2021mdetr}.
We choose FiLM \cite{perez2018film}, NSCL\cite{Mao2019NeuroSymbolic} and mDETR\cite{kamath2021mdetr} as category representatives. 

\textbf{VQA datasets.}
Datasets containing real images and human-written questions have been widely used to benchmark VQA models, \eg VQA \cite{antol2015vqa}, VQAv2 \cite{balanced_vqa_v2}, Visual 7w \cite{zhu2016visual7w}, VizWiz \cite{gurari2018vizwiz}, Visual Genome \cite{krishna2017visual}, COCO QA \cite{ren2015exploring}, etc. However, subsequent work has revealed the strong prior and bias in those datasets which might be exploited by models to correctly predict the answers without reasoning \cite{Niu2021CounterfactualVA, gupta2022swapmix, Niu_2021_CVPR, agrawal2016analyzing, balanced_vqa_v2, GQA-OOD, Kervadec2021HowTA}. Attempts to address this problem include better balancing datasets \cite{vqa-cp} and creating counterfactual examples \cite{Dancette2021BeyondQB, CSS}. To assess a model's true reasoning ability, the CLEVR dataset \cite{johnson2017clevr} proposes to generate complex multi-step questions on synthetic images,
which is then extended to various vision-and-language tasks \cite{liu2019clevr, kottur2019clevr, yi2019clevrer, zhang2019raven, bahdanau2019closure, salewski2022clevr, hong2021ptr}. 
The GQA dataset \cite{hudson2019gqa} extends CLEVR-style questions to real images. 
Our benchmark is distinct from existing ones because we introduce more complex visual scenes into CLEVR and provide controllability to study domain robustness on isolated factors.



\textbf{Domain shift in VQA.}
Domain shift is a long-standing challenge in computer vision, explore in prior works in domain adaptation \cite{ganin2015unsupervised, hoffman2018cycada, ganin2016domain, long2018conditional} and domain generalization \cite{qiao2020learning, li2019episodic}.
Recent works have focused on domain shifts in VQA. 
\cite{chao2018cross, xu2019open} improves model adaptation between datasets by feature learning. 
\cite{zhang2021domain} analyze domain shifts between nine popular VQA datasets and proposes an unsupervised method to bridge the gaps. \cite{Li_2021_ICCV} generalize symbolic reasoning from synthetic to real dataset. \cite{akula2021crossvqa} introduce a question-answer generation module that simulates the domain shifts. \cite{jiang2021x} propose a training scheme X-GGM to improve out-of-distribution generalization. \cite{bahdanau2019closure} assess models generalization on the CLOSURE of linguistic components. In contrast to prior works we study each of the different domain shift factors independently with our virtual benchmark.

\section{Super-CLEVR}

\subsection{Motivation: domain shift factors} \label{sec: motivation}


\textbf{Visual complexity.}
A major difference between different VQA datasets is visual complexity. 
For example, in the CLEVR dataset, objects are simple, atomic shapes 
while in real-world data, objects are more complex and have hierarchical parts.
While hard to quantify, visual complexity is related to various factors, such as object variety, object size, background, texture, lighting, occlusion, view point, etc. In our work, we control visual complexity by introducing more challenging objects that can have distinct attributes associated with their parts, and by optionally pasting various textures onto objects. Examples of generated images with different complexity levels are shown in \cref{fig:introduction}. 

\textbf{Question redundancy.}
Question redundancy refers to the amount of over-specified or redundant information in the question, which can be in the form of either \textit{attributes} or \textit{relationships}. For example, in \cref{fig:introduction}, ``what color is the large bus behind the cyan car'', \texttt{large} (\textit{attributes}) and \texttt{behind the cyan car} (\textit{relationship}) are redundant because there is only one bus in the image. As observed in linguistics and cognitive science \cite{ford1975elaboration, sonnenschein1985development, pechmann1989incremental, koolen2011effects}, human speakers may include over-specified information when identifying a target object, which has also been studied in referring expression generation \cite{mitchell2013generating}. For VQA, as analyzed in \cite{Li_2021_ICCV}, a significant difference between synthetic and real datasets is that real questions contain some redundant information, which sometimes is a distraction leading to model prediction errors. Therefore, in this work, we generate questions with different redundancy levels and study the effect of question redundancy on model behaviors.

\textbf{Concept distribution.}
The distributions of \textit{concept}, \ie \textit{objects} (\eg car) and \textit{attributes} (\eg large), are distinct across different VQA datasets. For example, while colors are well-balanced in CLEVR dataset, in the GQA dataset, the color distribution is long-tailed where ``white'' appears $>50$ times more frequently than ``gold''. Long-tailed distributions have been a challenge in many computer vision tasks \cite{liu2019large, zhang2017range, he2021re, van2018inaturalist, ju2021relational}.
In VQA, the long-tailed concept distribution not only hinders the learning of infrequent concepts due to few training samples, but also introduces strong biases and priors in the dataset that may mislead the models. For example, ``tennis'' is the correct answer to most questions with ``what sport is ...'' \cite{vqa-cp}. With strong priors in data, it is hard to assess the true reasoning capacity of current models. While previous works address this problem by carefully re-balancing datasets \cite{vqa-cp}, in our work, we controllably vary the concept distribution in our dataset and study model robustness to concept distribution shifts.

\textbf{Concept compositionality.}
Concept compositionality refers to how different concepts (shapes, attributes) compose and co-occur with each other, \eg roses are usually red while violets are usually blue \cite{GQA-OOD}. Concept compositionality can be viewed as a conditional concept distribution in the context of other concepts. Shifts in concept compositionality impede the generalization of VQA models. For example, the model may fail to recognize a \textit{green} banana because most bananas are \textit{yellow} in the training data  \cite{cadene2019rubi}. Previous works evaluate the out-of-distribution performance by collecting counterfactual testing examples \cite{Niu2021CounterfactualVA}. In our work, we control the compositionality of \textit{shapes} and \textit{colors} with an intuitive motivation: if, for example, in the training data, bicycles are red and cars are blue, will the models be able to recognize blue bicycles and red cars in testing?

\begin{figure}[t]
\begin{center}
\vspace{-0.5em}
\includegraphics[width=0.95\linewidth]{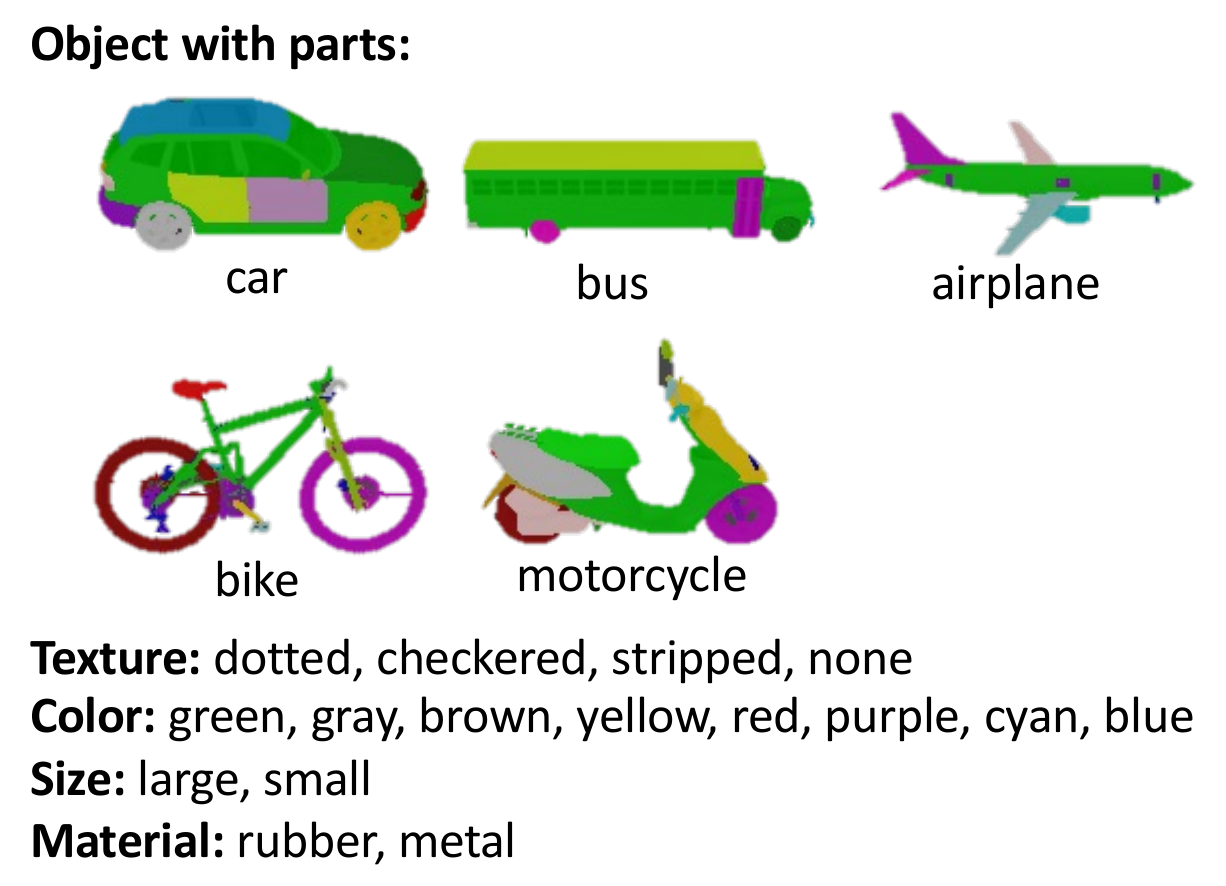}
\end{center}
    \vspace{-1.5em}
    \caption{Super-CLEVR contains 21 vehicle models belonging to 5 categories, with controllable attributes.}
\vspace{-1.5em}
\label{fig:dataset}
\end{figure}

\subsection{Dataset generation}
Super-CLEVR follows a similar data generation pipeline as CLEVR, but with more complex visual components and better control of domain gap factors. We describe the generation procedure below.


\textbf{Objects with parts.} To improve the visual complexity of CLEVR scenes, we replace the simple shapes (\eg,  \textit{cube}, \textit{sphere}) in CLEVR dataset with vehicles from UDA-Part dataset \cite{liu2022learning}. There are 21 vehicle models, belonging to 5 categories: \textit{car}, \textit{motorbike}, \textit{aeroplane}, \textit{bus}, and \textit{bicycle}. Each 3D model comes with part annotations, \eg, \textit{left front wheel} or \textit{left right door} for \textit{car}. Examples for the vehicle models are shown in \cref{fig:dataset}. We remove or merge small parts from the original annotations to avoid severe difficulty in visual understanding. The full object and parts list is in the supplementary material. 
 
\textbf{Attributes.} Besides the attributes in the original CLEVR dataset, \ie \textit{color}, \textit{material}, \textit{size}, we optionally add \textit{texture} as an additional attribute to increase visual complexity. Note that in order to enable part-based questions, the attributes (color or material) of object parts can be different from that of the object. For example, a blue car can have a red wheel or a green door. In this case, the attribute of the holistic object refers to the attribute of its main body (\eg the blue car has blue frame). 
 
\textbf{Scene rendering.} Following CLEVR, each scene contains 3 to 10 objects. The objects are placed onto the ground plane with random position and orientation. When placing the objects, we ensure that the objects do not overlap with each other and we avoid severe occlusion by thresholding the number of visible pixels for each object. Random jitters are added to lamp and camera positions. When rendering, we also save the ground-truth bounding boxes and segmentation masks for each of the objects and their parts, which are required when training some of the models.

\textbf{Question generation.} Super-CLEVR follows similar question generation pipeline in CLEVR, which instantiates question templates using the underlying reasoning program that can be operated on the scene graph. For example, the program \texttt{select\_shape(truck) $\to$ query\_color($\cdot$)} can be instantiated as question ``what is the color of the truck''. Therefore, redundancy level of questions can be controlled by removing or adding redundant reasoning steps in the underlying reasoning program.

\subsection{Controlling the dataset} \label{sec:controlling}
To study domain generalization, we generate several variants of the dataset for each of the domain shift factors. The variants of the datasets serve as different data domains to test the model robustness. Here we describe the method for controllably generating the dataset variants.

\textbf{Visual complexity.}
We generate three variants of the dataset with different levels of visual complexity: \textit{easy}, \textit{mid} (middle) and \textit{hard}. The only difference between the 3 versions is visual complexity: for the easy version, objects with different sizes, colors and materials are placed into the scene; for the middle version, we choose 3 parts on each object that are visible and randomly change their attributes; for the hard version, we further add random textures to the objects and parts. An example of the 3 dataset versions can be found in \cref{fig:introduction}. Note that the scene layout and the questions are shared, so that the influence of visual complexity can be isolated and studied independently.

\textbf{Question redundancy.}
Three variants of the dataset with different redundancy levels are generated: \textit{rd-}, \textit{rd} (default), \textit{rd+}. By default (\textit{rd}), as in original CLEVR dataset, the questions contain some redundant \textit{attributes} resulting from random sampling, while all redundant \textit{relationships} are removed. In \textit{rd-}, we also remove all redundant attributes from the questions, leading to no redundancy in the questions. In \textit{rd+}, we add all possible attributes and relationships into the question, so that questions contain a high level of redundancy. For all the variants, the questions are ensured to be valid.

\textbf{Concept distribution.}
We generate three dataset variants with different concept distributions: \textit{bal} (balanced), \textit{slt} (slightly unbalanced) and \textit{long} (long-tail distributed). More specifically, we change the distribution of \textit{shapes}, \textit{colors} and \textit{materials} while the distribution of \textit{size} is kept fixed in order to keep visual complexity consistent, since objects with smaller sizes are visually harder to recognize. By default (\textit{bal}), the shapes and attributes are randomly sampled, leading to a balanced distribution. For \textit{slt} and \textit{long}, the concept distribution $\mathbf{d}$ is generated by $d_{i} = a^{-i}$, where $i$ is the index of the concept. $a$ is a hyper-parameter controlling the length of the tail. A larger $a$ leads to more imbalanced distribution and $a=1$ leads to flat distribution  (cf. \cref{fig:introduction}). For \textit{slt}, $a=1.3$; for \textit{long}, $a=2.0$.
In addition, to better analyze the performance on the frequent and rare concepts, we generate three variants for testing purpose only: \textit{head} (frequent concepts in the long-tail distribution), \textit{tail} (infrequent/rare concepts), and \textit{oppo} (opposite to the long-tail distribution). We test each model on those three variants to analyze the performance on concepts with different degrees of frequency. 

\textbf{Concept compositionality.}
We generate 3 versions of the dataset, \textit{co-0}, \textit{co-1} and \textit{co-2}, with different compositions of the 21 \textit{shapes} (from 5 categories) and the 8 \textit{colors}. The compositionality of the dataset is controlled with the co-distribution matrix $M \in \mathbb{R}^{21 \times 8}$, where each entry $M_{ij}$ is the probability an object of the $i$-th shape has the $j$-th color. Entries in each row of $M$ sum up to $1$. 
In the version \textit{co-0}, $M$ is a flat matrix so that the shapes and colors are randomly composed. In \textit{co-1}, each shape in one category has a different color distribution, \eg truck and sedan, while shapes from different categories may share the same color distribution \eg sedan and airliner. Oppositely, in \textit{co-2}, we make the shapes in same category have the same color distribution, whiles shapes from different categories have different distributions. The motivation is that since shapes from the same category are visually similar, the difference in \textit{co-1} and \textit{co-2} will help analyze the difference in model predictions on visually similar objects and dissimilar objects when composed with different color distributions. 


\textbf{Dataset Statistics.}
Every dataset variant contains 30k images, including 20k for training, 5k for validation and 5k for testing. Each image is paired with 10 object-based and 10 part-based questions. By \textit{default}, the dataset refers to the version with \textit{mid} visual complexity level (\ie objects are untextured and has up to 3 parts with distinct attributes), \textit{rd} redundancy level, balanced (\textit{bal}) concept distribution and random (\textit{co-0}) compositionality. More dataset statistics are in supplementary materials.

\section{Evaluated methods}

\subsection{Simple baselines}

\textbf{Random, Majority.} These simple baselines pick a random or the most frequent answer for each question type in the training set as the predicted answer.

\vspace{-0.2em}
\textbf{LSTM.} This question-only baseline encodes the question word embeddings with LSTM \cite{hochreiter1997long} and predicts answer with an MLP on top of the final hidden states of the LSTM.

\vspace{-0.2em}
\textbf{CNN+LSTM.} The image is represented with features extracted by CNN and question is encoded by LSTM. An MLP predicts answer scores based on the concatenation of image and question features.

\subsection{Existing models}

\vspace{-0.2em}
\textbf{FiLM.} We choose \textit{Feature-wise Linear Modulation} \cite{perez2018film} as a representative of classic two-stream feature merging methods. The question features extracted with GRU \cite{cho2014learning} and image features extracted with CNN are fused with the proposed FiLM module.

\vspace{-0.2em}
\textbf{mDETR.} mDETR \cite{kamath2021mdetr} is a transformer-based detector trained to detects objects in an image conditioned on a text query. The model is pretrained with 1.3M image and text pairs and can be finetuned for various downstream tasks like referring expression understanding or VQA. 

\vspace{-0.2em}
\textbf{NSCL.} The \textit{Neuro-Symbolic Concept Learner} \cite{Mao2019NeuroSymbolic} is a representative neural symbolic method. NSCL executes neural modules on the scene representation based on the reasoning program, during which the modules learns embeddings of each concept with the answer supervision. 

\vspace{-0.2em}
\textbf{NSVQA.} \textit{Neural-Symbolic VQA} \cite{nsvqa} is a neural symbolic method composed of three components: A scene parser (Mask-RCNN \cite{he2017mask}) that segments an input image and recovers a structural scene representation, a question parser that converts a question from natural language into a program and a program executor that runs the program on the structural scene representation to obtain the answer. Notably, compared to NSCL, the individual components of NSVQA can be learned separately, hence, for example the scene parser can be learned from data that does not necessarily have Visual-Question annotations.

\subsection{Probabilistic NSVQA (P-NSVQA)}
Since the program executor in NSVQA is a collection of deterministic, generic functional modules, it can be augmented with a probabilistic reasoning process that takes into account the confidence of the predictions of the scene parser. This allows the model to execute the program that has the largest joint likelihood, instead of only taking the maximal likelihood execution at each step of the program. The experiment results demonstrate a significant performance improvement of this probabilistic approach over the deterministic NSVQA model proposed in \cite{nsvqa}. 

In particular, we interpret the confidence of the Mask-RCNN output as a likelihood function for all detected object classes $p_{object}$ and their attributes $p_{att}$. Moreover, we define a likelihood $p_{spatial}$ for the spatial relations between objects (behind, in front, left, right) that is proportional to the distance between the centers of two bounding boxes. Given a reasoning program containing multiple reasoning steps, we execute each step based on the scene parsing likelihood and produce an step-wise output with confidence. Finally, we use a factorized model, multiplying the output for all the steps to get the final answer prediction.
We refer readers to the Appendix for more details.


\subsection{Implementation details}

Training mDETR requires ground-truth grounding of question tokens to image regions, which is available in Super-CLEVR. 
NSCL requires bounding box of objects, which can predicted using a trained Faster RCNN, and the reasoning program, which can be parsed using a trained parser. Similarly, ground-truth programs are used for training NSVQA and P-NSVQA. Note that we empirically find that the question-to-program parsing is a relatively easy task ($>99\%$ accuracy using a simple LSTM), so we focus more on models' reasoning ability in our analysis.

Unless specified, the models are trained with default setting as in the official implementation.
FiLM is trained for 100k iterations with batch size 256. mDETR is trained for 30 epochs with batch size 64 using 2 GPUs for both the grounding stage and the answer classification stage. NSCL is trained for 80 epochs with batch size 32. 
For NSVQA and P-NSVQA, we first train the object parser (Mask RCNN \cite{he2017mask}) for 30k iterations with batch size 16, then train the attribute extraction model (using the Res50 backbone) for 100 epochs with batch size 64. For P-NSVQA, when counting the objects or determining whether objects exist in the scene, we use a threshold (0.7) to obtain the final selected objects. Early stopping is used based on validation accuracy. All the models are trained with 200k questions. We repeat experiments on \textit{default} split for 3 times with different random seeds and get \textit{std} of $\pm 0.10$ (P-NSVQA) and $\pm 0.40$ (NSVQA), showing the statistical significance of our results, then we only run other experiments once.

\section{Results and analysis}
In this section, we first show evaluation results for in-domain setting, then provide results and analysis on out-of-domain evaluation. Finally, we describe additional studies and future works.

\subsection{In-domain results}

\begin{figure}[h]
\begin{center}
\vspace{-0.5em}
\includegraphics[width=1.0\linewidth]{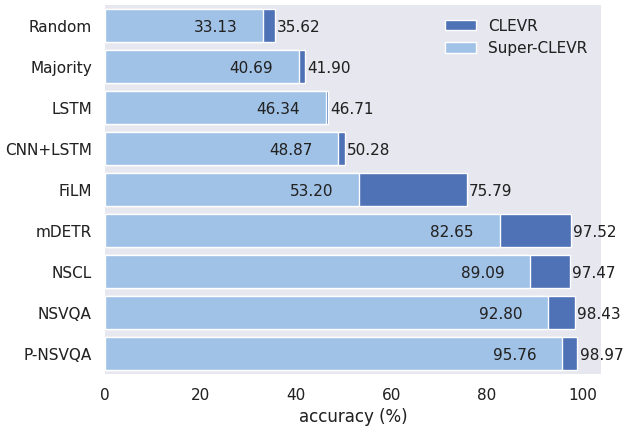}
\end{center}
    \vspace{-1.5em}
    \caption{Comparison of models' accuracy on Super-CLEVR and the original CLEVR dataset.}
\vspace{-0.5em}
\label{fig:baseline}
\end{figure}

In-domain evaluation refers to the setting where the training and testing data come from the same domain (the \textit{default} dataset variant in this case). We compare the in-domain results on Super-CLEVR and CLEVR. 
The results are shown in \cref{fig:baseline}.

 
For all the models, the performance is lower on Super-CLEVR than CLEVR, suggesting that Super-CLEVR is a more challenging benchmark. The scenes with vehicles are much harder visually for the models to understand compared with the simpler shapes from CLEVR. Note that the performance gap on two datasets for simple baselines (Random, Majority, LSTM, CNN+LSTM) is smaller than for the other better models. This is because Super-CLEVR contains more object types, and therefore the performance of simply guessing is lower than on CLEVR. 

When comparing the performance of different models, we find that the neural modular methods, \ie NSCL, NSVQA, P-NSVQA, perform much better than non-modular ones. This is not surprising given their nearly perfect performance on the original CLEVR dataset, which shows its strong ability to model synthetic images. 
The large-scale pretrained grounding model mDETR, which is a leading model on both real and synthetic images, also achieves good performance (82.7\%) on Super-CLEVR. The two-stream method FiLM does not achieve very strong performance (53.2\%), but is still much better than the other simple baselines.

Our proposed P-NSVQA outperforms all the other models. In particular, on Super-CLEVR, it outperforms its deterministic counterpart, NSVQA, by 2.96\%. This shows the advantage of taking into account the probabilities when the scenes are challenging thus the model's uncertainty of predictions can be utilized.

\subsection{Out-of-domain results} \label{sec: diagnose}

\begin{table*}[t]
\begin{center}
\vspace{-1.0em}
\resizebox{1.\linewidth}{!}{
\begin{tabular}{l|ccc|ccc|ccc|ccc|ccc}
\toprule
             & \multicolumn{3}{c|}{\textbf{FiLM}} & \multicolumn{3}{c|}{\textbf{mDETR}} & \multicolumn{3}{c|}{\textbf{NSCL}} & \multicolumn{3}{c|}{\textbf{NSVQA}} & \multicolumn{3}{c}{\textbf{Prob NSVQA}}\\
\bottomrule
\toprule
\multicolumn{16}{c}{\textbf{Visual Complexity}} \\ \midrule
       & easy     & mid       & hard       & easy      & mid       & hard       & easy     & mid       & hard    & easy      & mid       & hard       & easy     & mid       & hard   \\
\midrule
easy         & {\bf \underline{59.96}}    & \underline{53.95}     & \underline{50.66}      & {\bf \underline{93.36}}     & \underline{84.30}      & \underline{82.97}      & {\bf \underline{95.13}}    & \underline{92.31}     & \underline{90.81}   & {\bf \underline{95.19}} & \underline{94.19} & \underline{94.09} & {\bf \underline{96.76}} & \underline{95.98} & \underline{96.37}   \\
mid          & {\bf 57.41}    & 53.28     & 50.18      & {\bf 83.34}     & 82.36     & 81.27      & 84.5     & {\bf 89.10}      & 86.33   & 81.99 & 92.80 & {\bf 93.78} & 86.25 & {\bf 95.76} & 95.11   \\
hard         & {\bf 55.95}    & 53.11     & 50.47      & 79.71     & 79.94     & {\bf 80.71}      & 76.85    & 78.66     & {\bf 85.08}  & 73.11 & 79.71 & {\bf 92.65} & 79.81 & 86.47 & {\bf 95.36}   \\
\bottomrule
\toprule
\multicolumn{16}{c}{\textbf{Question Redundancy}} \\ \midrule
   & rd-      & rd        & rd+        & rd-       & rd        & rd+        & rd-      & rd        & rd+   & rd-       & rd        & rd+        & rd-      & rd        & rd+     \\
\midrule
rd-          & \underline{51.42}    & 52.54     & {\bf 53.51}      & {\bf \underline{83.94}}     & 80.37     & 66.28      & \underline{88.64}    & 88.82     & {\bf 90.33}   & {\bf \underline{92.95}}   & \underline{92.94}   & 92.67 & \underline{95.66}   & {\bf \underline{95.72}} & 95.43 \\
rd           & 50.39    & \underline{53.28}     & {\bf 54.78}      & {\bf 82.77}      & 82.36     & 70.36      & 88.45    & \underline{89.10}      & {\bf \underline{91.45}}  & 91.19   & {\bf 92.78} & 92.14 & 94.87 & {\bf \underline{95.72}} & 95.43    \\
rd+          & 46.14    & 52.30     & {\bf \underline{71.47}}      & 78.48     & \underline{84.05}     & {\bf \underline{90.42}}      & 87.94    & 88.34     & {\bf 91.16}  & 91.38   & 91.96 & {\bf \underline{92.80}} & 94.88 & 95.47 & {\bf \underline{95.72}}  \\


\bottomrule
\toprule
\multicolumn{16}{c}{\textbf{Concept Distribution}} \\ \midrule
 & bal      & slt       & long       & bal       & slt       & long       & bal      & slt       & long     & bal       & slt       & long       & bal      & slt       & long  \\
\midrule
bal     & \underline{50.47}    & 53.04           & {\bf 54.35}      & {\bf \underline{80.71}}     & 75.79          & 74.54      & {\bf 85.08}    & 83.79          & 75.10   & {\bf 92.65} & 90.82 & 83.74 & {\bf 95.36} &  94.89 & 89.88    \\
long    & 49.43    & \underline{54.75}           & {\bf \underline{62.96}}      & 79.06     & \underline{80.29}           & {\bf \underline{90.66}}      & \underline{85.33}    & \underline{89.42}          & {\bf \underline{91.10}} & \underline{92.73} & {\bf \underline{93.38}} & \underline{92.53} & \underline{96.31} & {\bf \underline{96.32}} & \underline{95.25}     \\ \midrule
head         & 48.60     & \underline{58.06}          & {\bf \underline{61.60}}       & 80.75     & \underline{79.60}          & {\bf \underline{87.46}}      & 84.58    & \underline{88.39}          & {\bf \underline{90.19}}    & \underline{93.87} & {\bf \underline{94.82}} & \underline{92.48} & \underline{96.42} & {\bf \underline{96.80}} & \underline{95.92}  \\
tail         & {\bf \underline{51.80}}     & 48.70          & 50.08      & {\bf \underline{81.50}}      & 70.88          & 60.94      & {\bf \underline{86.10}}     & 80.27          & 60.55    &  {\bf 90.26} & 89.20 & 75.32 & {\bf 94.08} & 93.20 & 82.68 \\
oppo        & {\bf 49.06}    & 48.93           & 46.68      & {\bf 79.13}     & 68.37          & 56.98      & {\bf 85.07}    & 77.86          & 55.14   &  {\bf 91.22} & 88.65 & 71.32 & {\bf 95.76} & 94.09 & 79.74 \\
\bottomrule
\toprule
\multicolumn{16}{c}{\textbf{Concept Compositionality}} \\ \midrule
  & co-0      & co-1       & co-2       & co-0       & co-1       & co-2       & co-0      & co-1       & co-2   & co-0       & co-1       & co-2       & co-0      & co-1       & co-2     \\
\midrule
co-0          & \underline{53.28}    & {\bf 57.00}        & 56.1       & {\bf \underline{83.36}}     & 77.03     & 82.43      & {\bf \underline{89.1}}     & 82.52     & 83.77   & {\bf \underline{92.80}} & \underline{90.11} & 91.59 & {\bf \underline{95.76}} & 94.02 & 95.12   \\
co-1          & 52.41    & {\bf \underline{60.57}}     & 56.67      & 79.46     & \underline{82.45}     & {\bf 83.93}      & 78.89    & {\bf \underline{87.18}}     & 84.2   & 78.74 & 89.99 & {\bf 90.67} & 87.12 & \underline{94.53} & {\bf 94.78}    \\
co-2         & 52.96    & 57.37     & {\bf \underline{60.53}}      & 80.03     & 77.41     & {\bf \underline{87.24}}      & 78.40     & 81.55     & {\bf \underline{88.84}}   &  77.85 & 89.28 & {\bf \underline{92.23}} & 87.19 & 93.49 & {\bf \underline{95.61}}  \\
\bottomrule
\end{tabular}}
\vspace{-0.5em}
\caption{Accuracy of models trained and tested on different domains. 
Column headings indicate \textit{training} settings, while rows indicate the dataset variant for \textit{testing}.
The best performance in each row (\ie the best training setting) is marked in \textbf{bold} and best performance in each column (\ie the best testing setting) is \underline{underlined}. Description for different splits is in \cref{sec:controlling} and analysis is in \cref{sec: diagnose}. }
\label{tab: results}
\end{center}
\vspace{-1.5em}
\end{table*}

In this section, we train and test the five models (FiLM, mDETR, NSCL, NSVQA and P-NSVQA) on different dataset variants, and diagnose their domain robustness
on each of the four domain shift factors. Please refer to \cref{sec:controlling} for a description of different variants. 
The validation accuracy is used for analysis here and the results are shown in \cref{tab: results}.

All the methods suffer from domain shifts. The results show that the best performance mostly occurs in situations where the model is tested on the same dataset variant as it is trained on, \ie the bold or underlined numbers fall mostly on the diagonals in \cref{tab: results}.

We compare the domain robustness of the five models by measuring the relative performance decrease when the testing data differs from the training data, \ie smaller performance drop on different testing domains means better robustness. Based on this intuition, for easier understanding of \cref{tab: results}, we propose a measurement metric for domain robustness named \textbf{\textit{Relative Degrade (RD)}} to better analyze the results.
We define \textit{Relative Degrade} as the the percentage of accuracy decrease when the model is tested under a domain shift, \ie the accuracy drop divided by the in-domain accuracy.  Specifically, if a model gets accuracy $a$ under in-domain testing (\ie testing with the same dataset variant as training) and accuracy $b$ under out-of-domain testing (\ie testing with a different dataset variant from training), then $RD = (a-b)/a$. Since we train each model on three data variants, the $RD$'s for the three models are averaged to measure its domain robustness.\footnote{For concept distributions, we compute relative degrade with a slight change: we compute the accuracy drop from \textit{head} to \textit{tail} and the drop from \textit{long} to \textit{oppo}, take their average, and divide by the accuracy on \textit{bal}.} 





\begin{table}[h]
\begin{center}
\begin{tabular}{lrrrr}
\toprule
 &
  \multicolumn{1}{l}{\textbf{Visual}} &
  \multicolumn{1}{l}{\textbf{Redund.}} &
  \multicolumn{1}{l}{\textbf{Dist.}} &
  \multicolumn{1}{l}{\textbf{Comp.}} \\ \midrule
\textbf{FiLM}  & {\bf 4.03}  & 21.33 & 28.46 & 9.04  \\
\textbf{mDETR} & 9.81  & 19.05 & 36.34 & 9.45  \\
\textbf{NSCL}  & 15.57 & 0.92  & 37.44 & 15.40\\ 
\textbf{NSVQA} & 17.48 & 1.72 & 20.92 & 11.44 \\ \midrule
\textbf{Prob NSVQA} & 12.88 & {\bf 0.84} & {\bf 13.72} & {\bf 7.00} \\
\bottomrule
\end{tabular}
\end{center}
\vspace{-1.5em}
\caption{\textit{Relative Degrade} under domain shifts, \ie the percentage of accuracy decrease when the model is tested with domain that differs with training. Lower \textit{RD} means better robustness.}
\label{tab: degrade}
\vspace{-1.0em}
\end{table}

\cref{tab: degrade} shows the \textit{Relative Degrade} of the five models on the four factors. We see that P-NSVQA outperforms other models by a significant margin on three of the four factors, indicating that it has better overall domain robustness. In the following, we take a closer look at the results on each of the factors separately, to diagnose the influence of different model designs.

\textbf{Question redundancy.}
Neural modular methods are much more robust to question redundancy shifts than non-modular ones. The relative degrades for modular methods are less than $2\%$, while  one-modular ones degrade for around $20\%$. Due to the step-by-step design of the reasoning in modular methods, each reasoning step is independent of the others so that the models are less likely to learn the spurious correlation between question and answers. Therefore the modular methods are less vulnerable to change in question/program length.

\textbf{Visual complexity.} 
Different from our findings on question redundancy, for domain shifts in visual complexity, non-modular methods are more robust compared to modular ones. As shown in \cref{tab: degrade}, while FiLM and mDETR gets less than $10\%$ degrade, NSCL and (P-)NSVQA degrade for more than $12\%$. The reason might be that the simple reasoning modules in modular methods can not process the visual signals as well as the dense non-modular models.

Comparing P-NSVQA with NSVQA, we find that injecting probability into deterministic symbolic reasoning greatly improves the robustness on visual complexity ($4.04\%$ decrease in \textit{RD}). This suggests that some errors in visual understanding can be corrected and recovered by taking into account the uncertainty of visual parsing and combining the results of each reasoning step with probability.


\textbf{Concept distribution.} 
While all the four existing models suffer a lot (larger than $20\%$ \textit{RD}) on domain shifts in concept distribution, we see that the symbolic method NSVQA is better than the other three (by more than $7.5\%$). With the disentangled reasoning and visual understanding components in NSVQA, the distribution priors in the images and the programs/answers cannot intertwine with each other, which prevent the model heavily relying on the priors. With uncertainty, we can further boost the robustness of NSVQA with a large margin (from $21\%$ to $14\%$ \textit{RD}). 


Moreover, the head-tail results suggests that the overall accuracy, which is commonly used to measure VQA performance, should be taken with cautious. When the testing split is imbalanced, the seemingly high accuracy is misleading because the head concepts dominates the testing while the tail ones are not well-reflected. For example, for NSCL, although it gets high accuracy (91\%) on the long-tailed data, its performance is only 60.6\% on the tail concepts. In real-world datasets, the data are usually not well-balanced, which suggests the value of synthetic testing.

\textbf{Concept compositionality.} 
Comparing the existing methods, we find that the non-modular methods seems to be more robust than modular methods NSCL or NSVQA. However, with uncertainty, P-NSVQA improves the result of NSVQA, which even outperforms the non-modular methods. This suggest the large potential of better robustness of modular methods by improving current models.

In summary, while non-modular methods are more robust to visual complexity shifts, the modular symbolic methods (improved with uncertainty) are more robust on the other three factors. By disentangling reasoning with visual understanding, separately executing every each reasoning step then merging the results of the steps using probabilities based on uncertainty, our P-NSVQA outperforms all the existing models in question redundancy, concept distribution and compositionality. Therefore, we suggest that symbolic reasoning with uncertainty leads to strong VQA models that are robust to domain shifts.

\subsection{More analysis and future work}
\textbf{Synthetic-to-real transfer.} We provide an additional proof-of-concept study to show that the findings drawn from Super-CLEVR dataset can transfer to real datasets. In the following experiments, we show our finding that neuro-symbolic methods (NSCL, NSVQA, P-NSVQA) are more robust than mDETR on question redundancy also holds true on the real GQA dataset \cite{hudson2019gqa}. More precisely, we progressively removed the redundant operations from the reasoning program in GQA testdev split, and then regenerated questions using a program-to-question generator. Using the change of models' testing accuracy as the redundant operations are removed, we can evaluate the models’ robustness towards question redundancy. The results are show in \cref{tab: GQA}.\footnote{For implementation of NSCL on a real-word dataset, we use the model in \cite{Li_2021_ICCV} (the version without calibration). The model accuracies on the original not-perturbed GQA testdev split are as following: mDETR (61.67\%), NSCL (56.13\%), NSVQA (39.58\%), P-NSVQA (39.66\%).} We observe that the performance drop of mDETR is much larger than neuro-symbolic methods as the redundant information is progressively removed, which indicates that symoblic methods have better question redundancy than mDETR on GQA dataset. This is consistent with our findings on Super-CLEVR. 

\begin{table}[h]
\begin{center}
\resizebox{1.\linewidth}{!}{
\begin{tabular}{lrrrrrr}
\toprule
\textbf{}   & \textbf{0\%} & \textbf{14\%} & \textbf{32\%} & \textbf{70\%} & \textbf{91\%} & \textbf{100\%} \\ \midrule
\textbf{mDETR} & 0            & -4.82         & -8.46         & -13.16        & -13.88        & -14.56        \\
\textbf{NSCL}  & 0            & -0.14         & -0.34         & -1.09         & -1.71         & -2.59          \\
\textbf{NSVQA}   & 0   & -3.47 & -4.80 & -7.01 & -7.02  & -7.02 \\
\textbf{P-NSVQA} & 0   & -1.93 & -3.15 & -5.73 & -5.91  & -5.78 \\
\bottomrule
\end{tabular}}
\end{center}
\vspace{-1.8em}
\caption{Accuracy drop on the GQA dataset when redundant information is progressively removed.}
\label{tab: GQA}
\vspace{-1.0em}
\end{table}


\textbf{Reasoning with part-object hierarchies.}
In addition to evaluating domain generalization, Super-CLEVR can be extended for broader purposes, \eg part-based reasoning. We can ask questions like ``what is the color of the front wheel of the bike?'', ``what is the color of the vehicle that has a yellow wheel'', etc. Those questions require the model to correctly understanding the part-object hierarchy, which is an ability that current VQA models lack. 

\textbf{Limitations.}
The main limitations of our work lie in the synthetic nature of our dataset. 
Future efforts can be made in collecting better controlled and balanced real datasets for model diagnosis. We emphasize that the purpose of the dataset is for model diagnosis and that models should also be tested on real data.

\section{Conclusion}

We diagnose domain shifts in visual reasoning using a proposed virtual benchmark, Super-CLEVR, where distinct factors can be independently studied with controlled data generation. We evaluate four existing methods and show that all of them struggle with domain shifts, highlighting the importance of out-of-domain testing. Among the evaluated methods, neural modular methods are more robust towards question redundancy. In particular, NSVQA with disentangled perception and reasoning shows better robustness towards distribution and compositionality shifts. We further propose P-NSVQA, which improves NSVQA with uncertainty in the reasoning modules. We show that P-NSVQA outperforms all the existing methods in both in-domain testing and out-of-domain testing. With detailed analysis, our study suggests that disentangling reasoning and perception, combined with probabilistic uncertainty, form a strong VQA model
that is more robust to domain shifts. 
We hope our analysis may facilitate better understanding of strengths and weaknesses of VQA models and, more broadly, future work might explore using the Super-CLEVR benchmark for other tasks like part-based reasoning. 

\subsection*{Acknowledgements}
\noindent
We would like to thank Nils Holzenberger, Kate Sanders, Chenxi Liu, Zihao Xiao, Qing Liu, Reno Kriz, and David Etter for their helpful comments and suggestions, as well as the anonymous reviewers. Zhuowan Li is supported by ONR N00014-21-1-2812 and grant IAA80052272 from the Institute for Assured Autonomy at JHU. Elias Stengel-Eskin is supported by an NSF GRFP. A. Kortylewski acknowledges support via his Emmy Noether Research Group funded by the German Science Foundation (DFG) under Grant No.468670075.

\clearpage
\newpage
{\small
\bibliographystyle{ieee_fullname}
\bibliography{references}
}

\clearpage
\appendix
\onecolumn

\section{Dataset statistics}
\begin{figure}[h!]
\begin{center}
\includegraphics[width=0.85\linewidth]{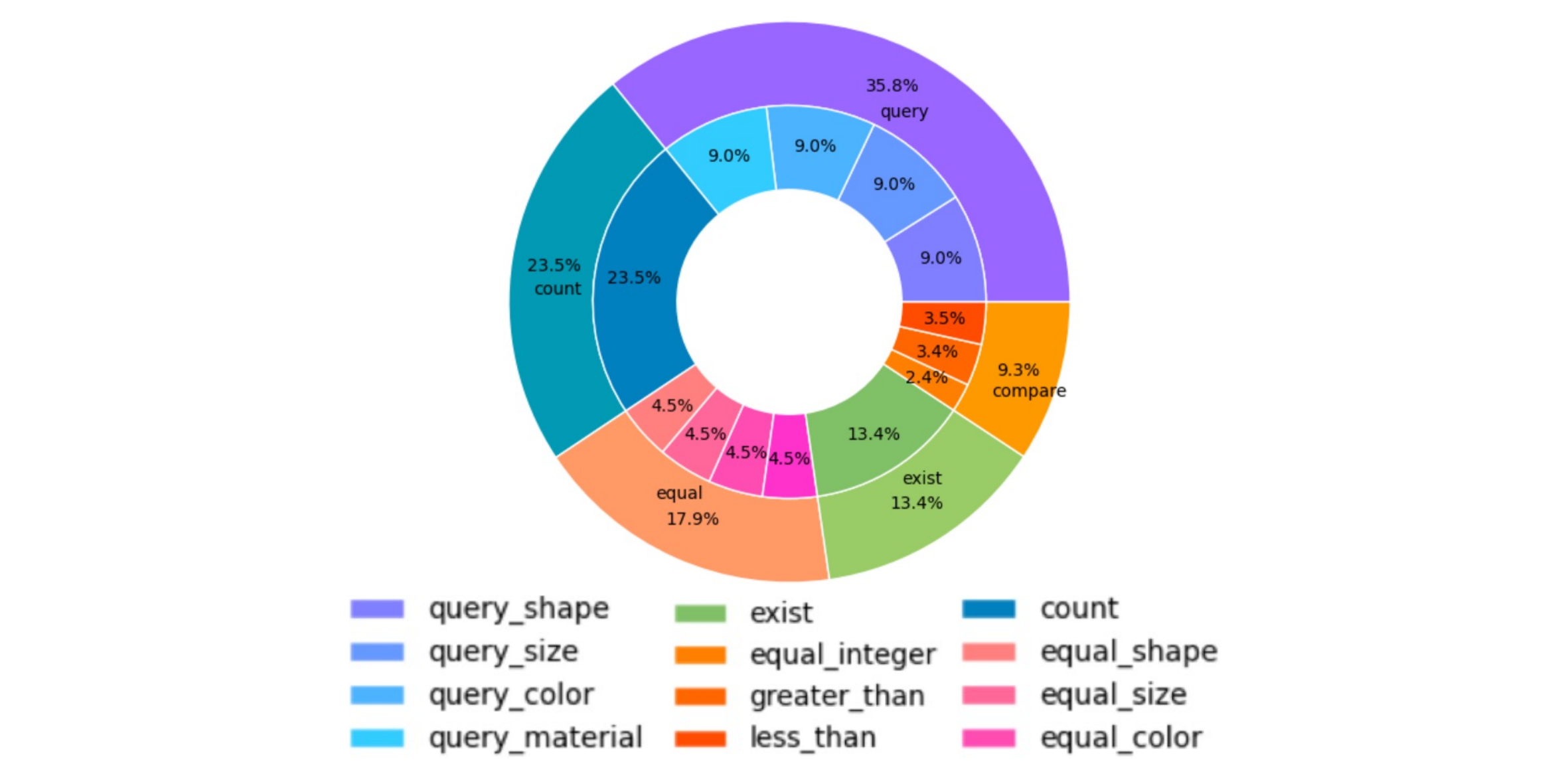}
\end{center}
\vspace{-1.5em}
   \caption{Distribution of question types in Super-CLEVR.}
\label{fig:dataset_dist}
\end{figure}

\cref{fig:dataset_dist} shows the distribution of question types in the Super-CLEVR dataset. The question type is determined by the type of the last operation in the question program.

\section{List of objects}

\begin{figure}[h!]
\begin{center}
\includegraphics[width=0.85\linewidth]{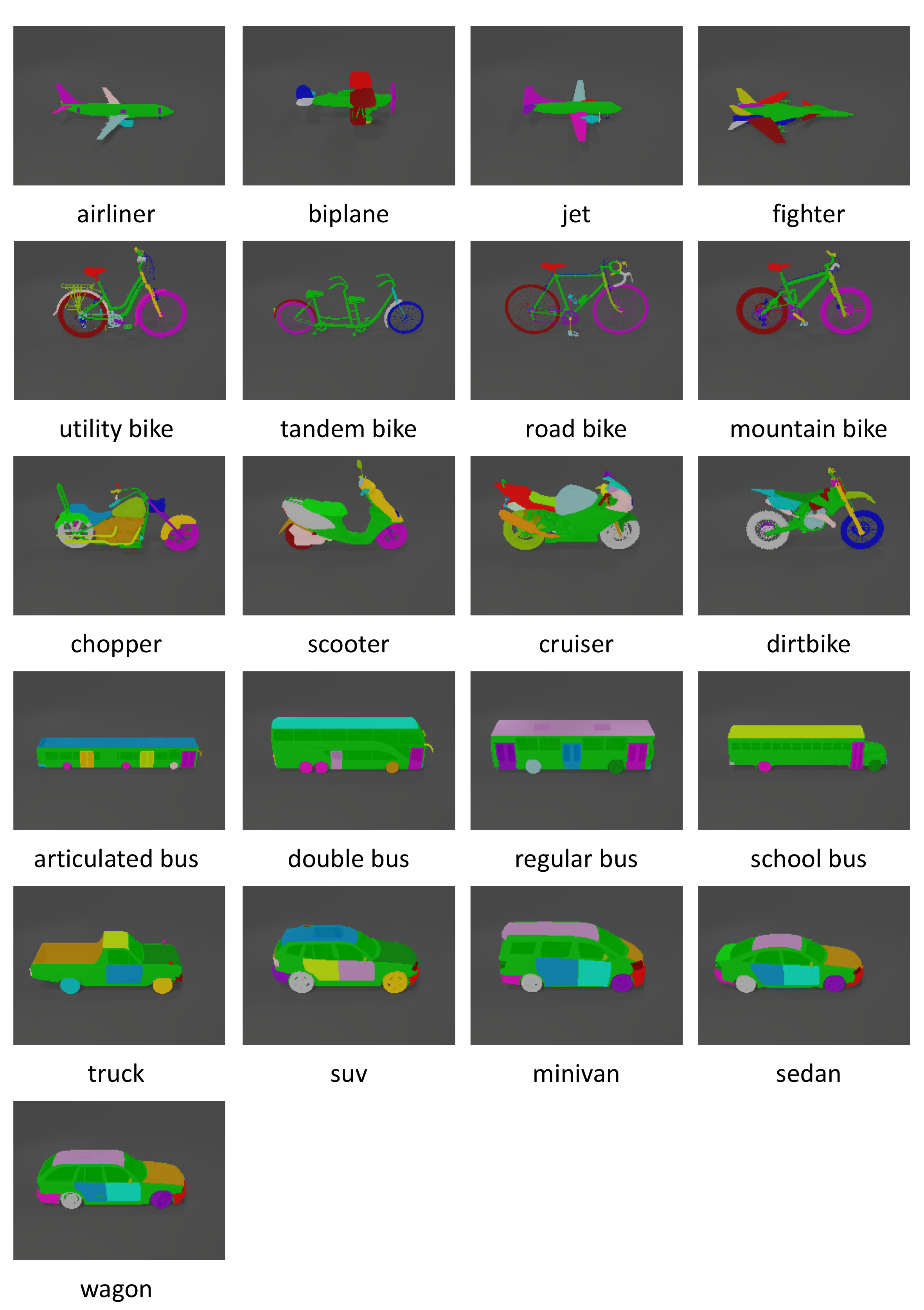}
\end{center}
   \caption{There are 21 objects belonging to 5 categories in the Super-CLEVR dataset.}
\label{fig:appendix_objects}
\end{figure}

Super-CLEVR contains 21 objects from 5 categories: airplane, bicycle, bus, car and motorcycle. They are shown in \cref{fig:appendix_objects} and \cref{tab: appendix_21shapes}. 

\begin{table}[h!]
\begin{center}
\begin{tabular}{cc}
\toprule
\multicolumn{1}{c}{\textbf{catetory}} & \multicolumn{1}{c}{\textbf{objects}}                 \\
\midrule
\textbf{airplane}                     & airliner, biplane, jet, fighter                      \\
\textbf{bicycle}                      & utility bike, tandem bike, road bike, mountain bike  \\
\textbf{bus}                          & articulated bus, double bus, regualr bus, school bus \\
\textbf{car}                          & truck, suv, minivan, sedan,wagon                     \\
\textbf{motorcycle}                   & chopper, scooter, cruiser, dirtbike                 \\
\bottomrule
\end{tabular}
\end{center}
\vspace{-1em}
\caption{Super-CLEVR dataset contains 21 objects from 5 categories.}
\label{tab: appendix_21shapes}   
\end{table}

\section{Dataset controlling} 
\begin{figure}[]
\vspace{-1.0em}
\begin{center}
\includegraphics[width=0.8\linewidth]{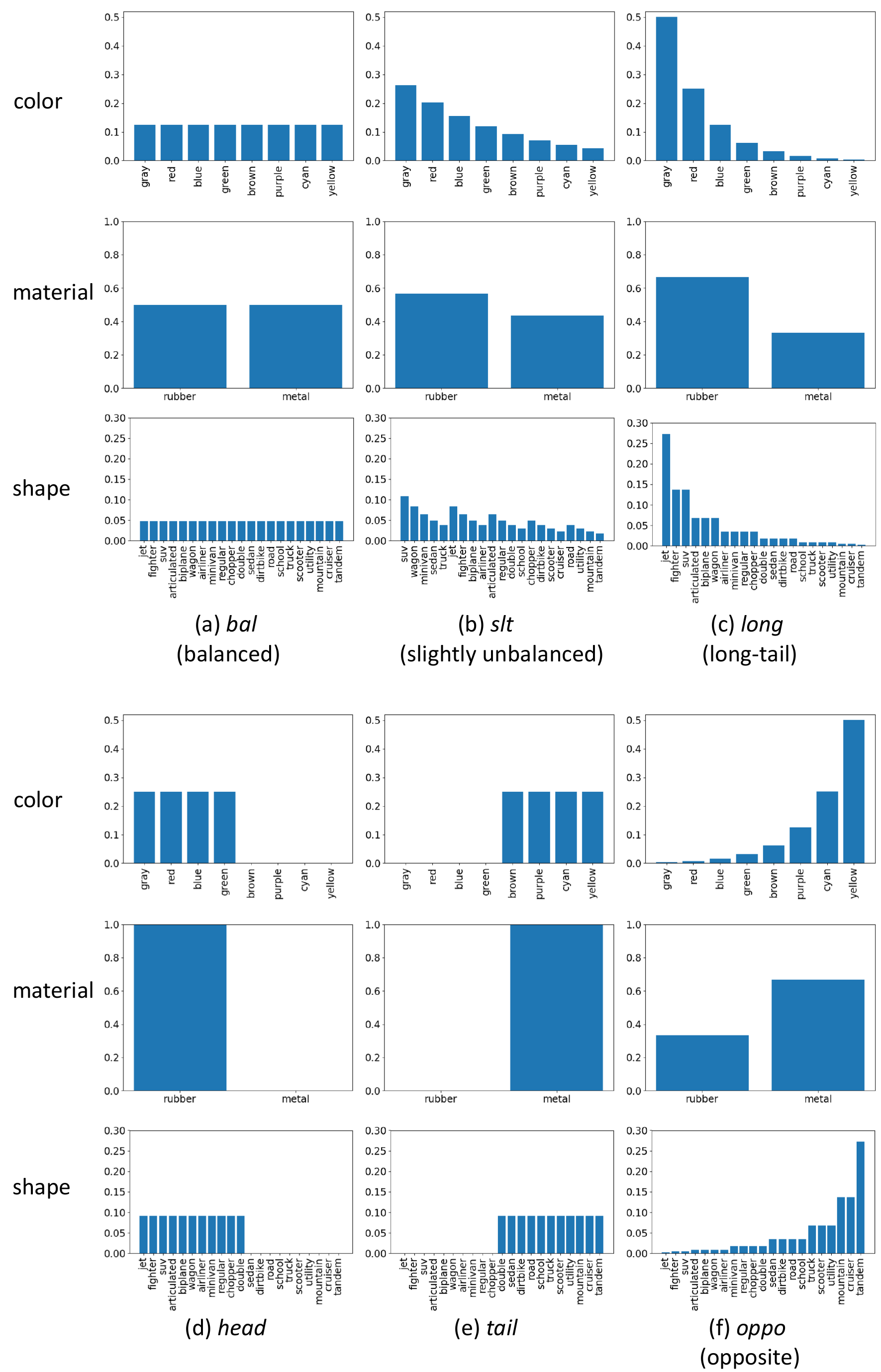}
\end{center}
\vspace{-1.5em}
\caption{Concept distribution for dataset variants \textit{bal}, \textit{slt}, \textit{long}, \textit{head}, \textit{tail} and \textit{oppo}.}
\label{fig:appendix_dist}
\end{figure}

\begin{figure}[]
\begin{center}
\includegraphics[width=0.95\linewidth]{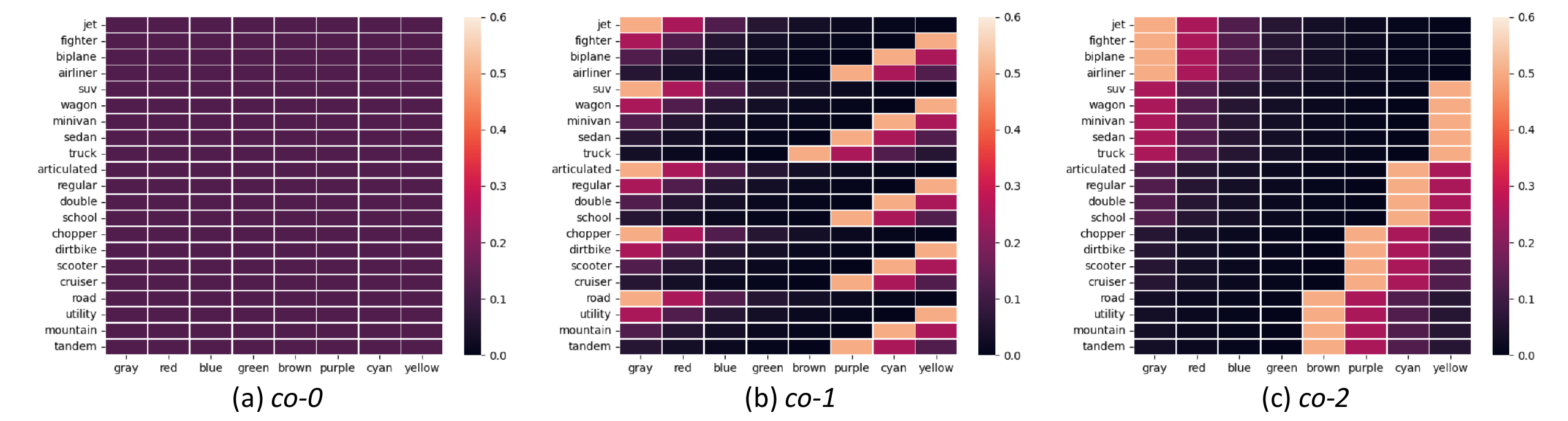}
\end{center}
   \caption{Concept co-distribution matrix $M$ for dataset variants \textit{co-0}, \textit{co-1} and \textit{co-2}.}
\label{fig:appendix_codist}
\end{figure}

\cref{fig:appendix_dist} shows the concept distribution for dataset variants \textit{bal}, \textit{slt} and \textit{long}; and variants \textit{head}, \textit{tail} and
\textit{oppo} for testing purpose. 
\cref{fig:appendix_codist} shows the concept co-distribution Matrix $M$ for controlling the concept compositionality (for variants \textit{co-0}, \textit{co-1} and \textit{co-2}).
The descriptions for the variants are in \cref{sec:controlling}.

\section{Definition of Relative Degrade}
Here we describe \textit{Relative Degrade} for each domain shift factors. We use $A^{i}_{j}$ to denote the accuracy of model trained using data variant $i$ and tested with data variant $j$. 

Visual complexity:
{\tiny
$$RD = Avg(\frac{A^{easy}_{easy} - A^{easy}_{mid}}{A^{easy}_{easy}}, \frac{A^{easy}_{easy} - A^{easy}_{hard}}{A^{easy}_{easy}}, \frac{A^{mid}_{mid} - A^{mid}_{easy}}{A^{mid}_{mid}}, \frac{A^{mid}_{mid} - A^{mid}_{hard}}{A^{mid}_{mid}}, \frac{A^{hard}_{hard} - A^{hard}_{easy}}{A^{hard}_{hard}}, \frac{A^{hard}_{hard} - A^{hard}_{mid}}{A^{hard}_{hard}})$$}

Question redundancy, 
{\tiny
$$RD = Avg(\frac{A^{rd\text{-}}_{rd\text{-}} - A^{rd\text{-}}_{rd}}{A^{rd\text{-}}_{rd\text{-}}}, \frac{A^{rd\text{-}}_{rd\text{-}} - A^{rd\text{-}}_{rd\text{+}}}{A^{rd\text{-}}_{rd\text{-}}}, \frac{A^{rd}_{rd} - A^{rd}_{rd\text{-}}}{A^{rd}_{rd}}, \frac{A^{rd}_{rd} - A^{rd}_{rd\text{+}}}{A^{rd}_{rd}}, \frac{A^{rd\text{+}}_{rd\text{+}} - A^{rd\text{+}}_{rd\text{-}}}{A^{rd\text{+}}_{rd\text{+}}}, \frac{A^{rd\text{+}}_{rd\text{+}} - A^{rd\text{+}}_{rd}}{A^{rd\text{+}}_{rd\text{+}}})$$}

Concept distribution,
{\tiny
$$RD = \frac{1}{3} \sum_{k \in S} \frac{(A^{k}_{head} - A^{k}_{tail}) + (A^{k}_{long} - A^{k}_{oppo})}{2 \cdot A^{k}_{k}}, S=\{bal, slt, long\}$$}

Concept compositionality,
{\tiny
$$RD = Avg(\frac{A^{co\text{-}0}_{co\text{-}0} - A^{co\text{-}0}_{co\text{-}1}}{A^{co\text{-}0}_{co\text{-}0}}, \frac{A^{co\text{-}0}_{co\text{-}0} - A^{co\text{-}0}_{co\text{-}2}}{A^{co\text{-}0}_{co\text{-}0}}, 
\frac{A^{co\text{-}1}_{co\text{-}1} - A^{co\text{-}1}_{co\text{-}0}}{A^{co\text{-}1}_{co\text{-}1}}, 
\frac{A^{co\text{-}1}_{co\text{-}1} - A^{co\text{-}1}_{co\text{-}2}}{A^{co\text{-}1}_{co\text{-}1}}, 
\frac{A^{co\text{-}2}_{co\text{-}2} - A^{co\text{-}2}_{co\text{-}0}}{A^{co\text{-}2}_{co\text{-}2}}, 
\frac{A^{co\text{-}2}_{co\text{-}2} - A^{co\text{-}2}_{co\text{-}1}}{A^{co\text{-}2}_{co\text{-}2}})$$
}

\vspace{1em}
\section{More details about P-NSVQA}

Given a image containing $n$ objects, we maintain a vector of probability $\mathbf{p}=[p^1, p^2, \ldots, p^n]$, where $p^k$ means the probability that object $k$ is selected. We update $\mathbf{p}$ when executing the reasoning operations step by step. In the following, we describe how to compute $\mathbf{p}$ for each kind of operations.

\begin{itemize}
    \item $scene$ \\
    Initialize all the values in $\mathbf{p}$ to 1.
    \item $filter_{identifier}[attribute]$ (\eg $filter_{color}[red]$) \\
    For object $k$, $$ p^k = p^k * P^k_{attribute} $$ Here $P^k_{attribute}$ is the probability of object $k$ having the $attribute$, which is predicted by the visual scene parsing model.

\item $relate\_{spacial}$ (including $relate\_{behind}$, $relate\_{front}$, $relate\_{right}$, $relate\_{left}$)

The output of the \textit{relate} operation is the probabilities of each object being on the $spacial$ side of the given object. For example, $relate\_{left}(i)$ computes the probabilities of objects to be on the left side of the given object $i$.
$$
p^{k}_{front} = \frac{1}{1+e^{-b[(y_k-y_i)+a]}}
$$

$$
p^{k}_{behind} = \frac{1}{1+e^{-b[(y_i-y_k)+a]}}
$$

$$
p^{k}_{right} = \frac{1}{1+e^{-b[(x_k-x_i)+a]}}
$$

$$
p^{k}_{left} = \frac{1}{1+e^{-b[(x_i-x_k)+a]}}
$$
Here $i$ is the input object and $(x_i, y_i)$ is the center of it. $a$ and $b$ are hyperparameters. In our experiments, we set $a=20$ and $b=0.02$. 

\item $same\_{color}$, $same\_{shape}$, $same\_{size}$, $same\_{material}$ \\
The \textit{same} operation returns the probabilities of each object having the same attribute as the given object $i$.
For example, for object $k$ and attribute \textit{color},

$$
p^k = cosine\_similarity (P^k_{color}, P^i_{color})
$$

\item $intersect$, $union$\\
Given two probability vectors $\mathbf{p_1}, \mathbf{p_2}$, we calculate their intersection or union:

$$Intersection: \mathbf{p} = \mathbf{p_1} \odot \mathbf{p_2}$$

$$Union: \mathbf{p} = 1 - (1- \mathbf{p_1}) \odot (1-\mathbf{p_2})$$
Here $\odot$ is the pointwise product.

\end{itemize}

\end{document}